\documentclass[conference]{IEEEtran}
\IEEEoverridecommandlockouts
\usepackage{cite}
\usepackage{amsmath,amssymb,amsfonts}
\usepackage{algorithmic}
\usepackage{graphicx}
\usepackage{textcomp}
\usepackage{xcolor}
\usepackage{hyperref}
\usepackage{subfig}
\def\BibTeX{{\rm B\kern-.05em{\sc i\kern-.025em b}\kern-.08em
    T\kern-.1667em\lower.7ex\hbox{E}\kern-.125emX}}
\begin{document}

\title{Image Compression And Actionable Intelligence With Deep Neural Networks}

\author{\IEEEauthorblockN{Matthew Ciolino}
\IEEEauthorblockA{\textit{PeopleTec Inc.} \\
Huntsville, AL, USA \\
matt.ciolino@peopletec.com}
}

\maketitle

\begin{abstract}
If a unit cannot receive intelligence from a source due to external factors, we consider them disadvantaged users. We categorize this as a preoccupied unit working on a low connectivity device on the edge. This case requires that we use a different approach to deliver intelligence, particularly satellite imagery information, than normally employed. To address this, we propose a survey of information reduction techniques to deliver the information from a satellite image in a smaller package. We investigate four techniques to aid in the reduction of delivered information: traditional image compression, neural network image compression, object detection image cutout, and image to caption. Each of these mechanisms have their benefits and tradeoffs when considered for a disadvantaged user.
\end{abstract}


\section{Introduction}
In a low connectivity environment, we need to be able to deliver as much value from the data we deliver as possible. To do so we must describe the benefits of sending alternative forms of information besides raw satellite imagery. This would depend on two things about the sent information, the size of the data and the effectiveness of the delivered information. In our past paper \cite{ciolino2022effectiveness}, we describe how different machine learning (ML) systems would impact the sensor to shooter timeline and how those techniques would improve the quality of information. We now seek to quantify the size of data reduction for various ML systems in addition to the value of information they supply. In this effort we can rank each method according to its ratio of impact versus data size. 

Various industries have data reduction techniques that lower infrastructure costs, allow more bandwidth, and improve the speed of communications. For example, Zoom will shrink your video, reduce the frame rate, or lower the bit rate of your video call to make sure your call is uninterrupted \cite{barolo_2020}. In addition to active compression, changing a file's format may also help in compression. While many file formats have been proposed and have been standardized for different forms of communication as discussed by Mahboob et al \cite{akila2018overview}, we shall look at compression techniques instead of a file format comparison.

\begin{figure*}[t]
    \centering
    \subfloat[\centering Autoencoder Architecture]{{\includegraphics[width=.45\textwidth]{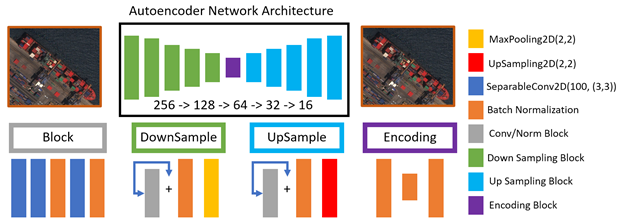} }}%
    \qquad
    \subfloat[\centering Image to Caption and Object Detection Image Compression]{{\includegraphics[width=.45\textwidth]{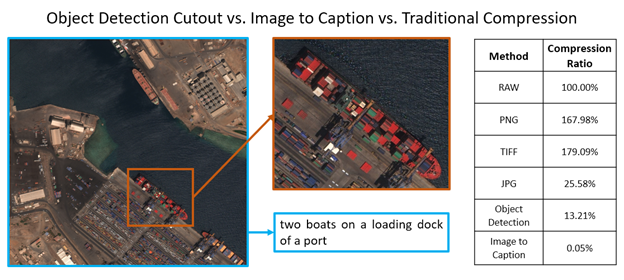} }}%
    \caption{Alternative Image Compression Methods}%
    \label{fig:example}
\end{figure*}

\subsection{Background}
Uthayakumar et al. \cite{jayasankar2021survey} surveyed the data compression techniques used by various data types. He shows a flow chart of those techniques for Text, Image, Audio, and Video compression. Our target for this paper is to investigate ML techniques for image compression. Hussain et al \cite{hussain2018image} describes various traditional image compression techniques including lossless methods: Arithmetic Coding \cite{clarke1995digital}, Huffman Coding \cite{huffman1952method}, Lossless predictive coding \cite{gonzalez1992digital} and lossy methods: Predictive Coding \cite{al2007hybrid}, Transform Coding \cite{clarke1985transform}, JPEG \cite{jtc2000inforamtion}, Vector Quantization \cite{gersho1982structure}. While these techniques have benefits and drawbacks they have been implemented in various database and communications fields such as television, satellite transmission, video conferencing, multispectral images, fingerprints, and medical images \cite{hussain2018image}. 

That leads us into this paper’s methods of using a neural network (NN) to reduce the size of delivered information. NNs can be considered a lossy compression method due to the reconstruction of an image from a compressed state. This can generally be called an Autoencoder (AE) \cite{ballard1983parallel} network where we compress an image to a representative state (embedding) and then expand it back into an image. The ratio of the size of the image to the size of the embedding is the compression ratio of the NN. Since the method is lossy, we can run common image metrics to compare the original image to the reconstructed image. Commonly used are peak-signal-to-noise ratio (PSNR) and structural similarity index measure (SSIM) \cite{hore2010image}. These metrics may be misleading as compared to no-reference metrics that have be subjectively constructed as researched by Chao et al \cite{ma2017learning}.  Regardless, compression ratio and PSNR are used to rank NN compression models. Comparisons between bitrate, the average number of bits needed to encode each image pixel information \cite{comparisonbits}, and performance is a heavily researched topic with Cheng et al \cite{cheng2018performance} \cite{cheng2018deep} \cite{cheng2019perceptual} discussing its effect across many different types of NNs including Generative Adversarial Networks (GAN) \cite{goodfellow2014generative}, Super Resolution (SR) \cite{ciolino2020training} \cite{yang2019deep}, and others \cite{smaller}.

While traditional compression methods and AE’s are intuitive to understand, other forms of data compression must be considered. Mainly two forms of alternative compression, image to caption and object detection. Image to caption \cite{noever2020discoverability} \cite{lu2017exploring} models are a combination of computer vision (CV) models and natural language processing (NLP) models. They ingest a picture and output a sentence describing the photo. If we can compress the information from a picture into a sentence in an impactful way, sending the photo would become superfluous. Alternatively, we might want to send just a portion of the total image. In this manner we would reduce the amount of data sent while still delivering the most important parts of the image. For example, if we are looking for tanks in satellite imagery, we could use an object detection model \cite{redmon2018yolov3} \cite{he2017mask} to give a bounding box around the objects of interest and only send those cutouts instead of the whole image. 

\subsection{Contributions}
In this paper we run an ablation of NN models to compare compression ratio to performance. We take an AE model and train at various embedding sizes. We then compare their compression ratio to both traditional compression techniques, such as JPEG \cite{gersho1982structure}, and alternative compression techniques like image to caption and object detection cutouts. We specifically focus on the xView Dataset \cite{lam2018xview}, a collection of over 1 million objects from WorldView-3 satellites at 0.3m resolution. We release the code for training the AE in colab \cite{ciolino_2022}.

\section{Experiments}

\subsection{Traditional Compression}

Various image compression techniques exist so we choose representative lossless and lossy methods to compare to each other. We gathered a RAW (.ARW) image and converted the image into a JPG, PNG, and a TIF. These are by far the most popular file formats for a machine learning dataset and will show the stark difference in compression ratios.

\subsection{Alternative Methods}

Complementing the traditional methods, we propose alternative methods including object detection image cutouts and image to caption. Both of these machine learning models, if they can function effectively, drastically reduce the size of the delivered information. Object detection will crop to the object of interest while image to caption will output a sentence describing the image. 

\subsubsection{Object Detection}

We use the xView validation dataset and run a pretrained YoloV3 \cite{glenn} \cite{redmon2018yolov3} to extract bounding box locations. We then crop the image to those bounding boxes and save the cutout to file. 

\subsubsection{Image to Caption}

We train on the Remote Sensing Image Caption Generation (RSCID) \cite{lu2017exploring} dataset, a collection of 10,000 images with 5 sentences describing each, and then run inference on sample xView images. The output is a sentence describing the scene. 

\subsection{Autoencoder}

To perform this ablation, we target a range of compression ratios for a convolutional based neural network. This network consists of residual connections between down sample and up sample blocks as shown in [Figure \ref{fig:example}A]. We resize the input images into 256x256 and then run them through the network. Through this we can see how compression ratio impacts the quality metrics including PSNR and SSIM. We run this analysis for block sizing one to five extracting the reduced image at 128, 64-, 32-, 16- and 8-pixel resolution along the way. We train at a 20\% validation split with a batch size of 25.

\section{Evaluation}

\subsection{Method Comparison}

 The comparison between traditional compression methods and the alternative methods comes down the quality of the output rather than the data reduction [Figure \ref{fig:example}B]. For example, lossy methods like JPG, object detection, image to caption, and an autoencoder lose information about the image. While lossless methods are great image quality, they take up a significant amount of data compared to the 25\% compression ratio of JPG. Alternative methods like object detection perform great in this regard since they aim to only transmit the important information, while potentially losing out on context information that might be lost. Image to caption makes up for this by trying to summarize the scene into a single sentence at a 0.05\% compression ratio. A combination of these methods, sent as metadata, may be the most effective way to transmit the most meaningful information besides the image itself.

\subsection{Autoencoder}

\begin{figure}[!t]
    \centering
    \includegraphics[width=\linewidth]{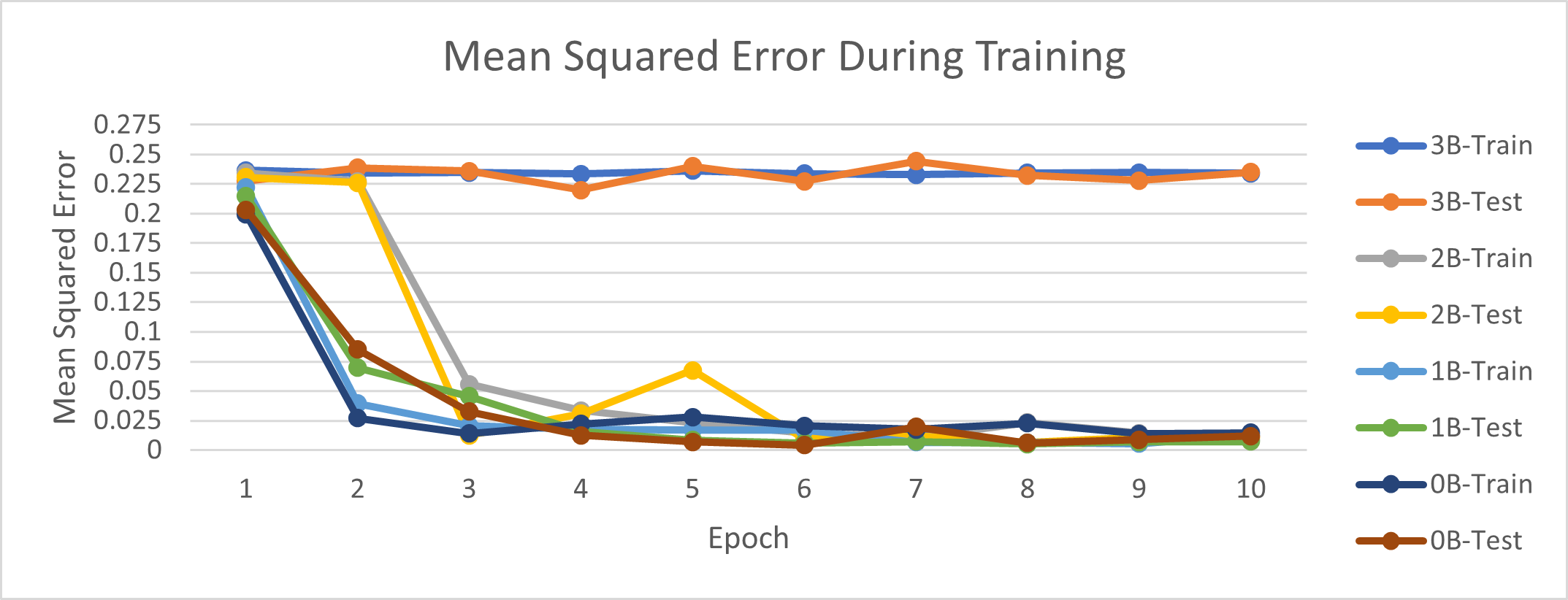}
    \caption{Autoencoder Training Results}
    \label{results}
\end{figure}

The autoencoder was able to perform well for such a small model (less than 1 million parameters). In the world of SR, anything above a 30 PSNR is considered remarkable. We were able to achieve lossy but good results for a compression ratio of 25\% (62 PSNR) and 6.25\% (26 PSNR). This means we can directly compare our AE to the JPG compression method detailed above. According to \cite{somwanshi2018modified}, JPG can be considered as having a PSNR of 40. Therefore, with the same compression ratio, our AE was able to achieve 43.14\% increases in PSNR as compared to JPG.  Mean Squared Error (MSE) loss during training is shown in [Figure \ref{results}] while the PSNR for various compression ratios is shown in [Table \ref{tab:my-table}]. All training runs with more than 3 blocks was not able to converge (Same as 3B-Train). 

% \usepackage{graphicx}
\begin{table}[]
\caption{Autoencoder Image Metrics vs Compression Ratio}
\label{tab:my-table}
\resizebox{\linewidth}{!}{%
\begin{tabular}{|c|c|c|c|c|c|c|}
\hline
\textbf{Blocks} & \textbf{PSNR Train} & \textbf{SSIM Train} & \textbf{PSNR Test} & \textbf{SSIM Test} & \textbf{Output Size} & \textbf{Compression} \\ \hline
0 & 63.9578 & 0.9869 & 39.7156 & 0.8342 & 256 & 100.00\% \\ \hline
1 & 62.4227 & 0.9778 & 51.0635 & 0.8426 & 128 & 25.00\% \\ \hline
2 & 26.6430 & 0.8612 & 20.3815 & 0.6586 & 64 & 6.25\% \\ \hline
3 & 6.7272 & 0.1863 & 6.2881 & 0.2731 & 32 & 1.56\% \\ \hline
4 & 6.4098 & 0.2159 & 7.0990 & 0.1720 & 16 & 0.39\% \\ \hline
5 & 7.0950 & 0.0005 & 5.9980 & 0.3817 & 8 & 0.10\% \\ \hline
\end{tabular}%
}
\end{table}

\section{Conclusion}
Various machine learning methods outperform common compression techniques. These applications of image compression can help a disadvantaged user get the data the matters to them the most in the right order. We propose a hierarchical transmission, where the smallest amount of impactfull data is sent first while the raw image is sent last. In this fashion, image to caption, object detection, and an autoencoder compressing an image can make an important step toward providing a user with the information they need. With this method, a disadvantaged user is not left waiting for a response but instead has actionable intelligence faster and in an effective format. 

\section*{Acknowledgment}
The authors would like to thank the PeopleTec Technical Fellows program for encouragement and project assistance. The views and conclusions contained in this paper are those of the authors and should not be interpreted as representing any funding agencies.

\bibliographystyle{./IEEEtran.bst}
\bibliography{./refs.bib}

\end{document}